\documentclass[conference]{IEEEtran}
\usepackage{times}
 
\usepackage[numbers]{natbib}
\usepackage{multicol}
\usepackage[bookmarks=true]{hyperref}

\usepackage{epsfig}
\usepackage{graphicx}
\usepackage{amssymb}
\usepackage{url}
\usepackage{mathtools}
\usepackage{bbm}
\usepackage{color}
\usepackage{booktabs}

\newcommand{\todo}[1]{{\color{white}{#1}}}
\newcommand{\act}{{\bf a}}
\newcommand{\acti}{a}
\newcommand{\obs}{\mathbf{I}}
\newcommand{\vel}{\mathbf{v}}
\newcommand{\at}{\act_t}
\newcommand{\ot}{\obs_t}
\newcommand{\as}{\act_s}
\newcommand{\os}{\obs_s}
\newcommand{\disc}{\gamma}

\DeclareMathOperator*{\argmax}{arg\,max}

\renewcommand{\baselinestretch}{0.95}

\begin{document}

\title{CAD$^2$RL: Real Single-Image Flight Without a Single Real Image\vspace{-0.05in}}

\author{\authorblockN{Fereshteh Sadeghi}
\authorblockA{University of Washington\\
fsadeghi@cs.washington.edu}
\and
\authorblockN{Sergey Levine}
\authorblockA{University of California, Berkeley\\
svlevine@eecs.berkeley.edu}
}

\maketitle

\begin{figure}
\vspace{-0.3in}
\begin{minipage}{1\textwidth}
        \centering
          \includegraphics[width=0.999\textwidth]{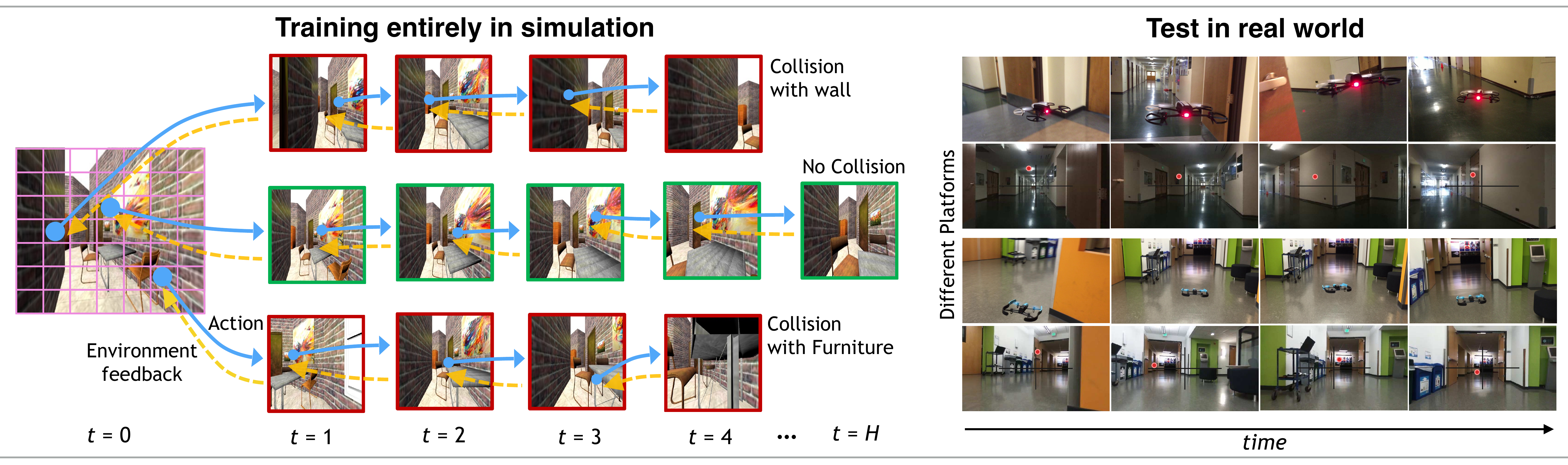}
          \vspace{-0.25in}
          \caption{We propose the {\bf C}ollision {\bf A}voidance via {\bf D}eep {\bf R}einforcement {\bf L}earning algorithm for indoor flight which is entirely trained in a simulated {\bf CAD} environment. Left: CAD$^2$RL uses \emph{single image} inputs from a monocular camera, is exclusively trained in simulation, and does not see any real images at training time. Training is performed using a Monte Carlo policy evaluation method, which performs rollouts for multiple actions from each initial state and trains a deep network to predict long-horizon collision probabilities of each action. Right: CAD$^2$RL generalizes to real indoor flight.}
          \vspace{-0.27in}
        \label{fig:teaser}
\end{minipage}
\end{figure}

\begin{abstract}
Deep reinforcement learning has emerged as a promising and powerful technique for automatically acquiring control policies that can process raw sensory inputs, such as images, and perform complex behaviors. However, extending deep RL to real-world robotic tasks has proven challenging, particularly in safety-critical domains such as autonomous flight, where a trial-and-error learning process is often impractical. In this paper, we explore the following question: can we train vision-based navigation policies entirely in simulation, and then transfer them into the real world to achieve real-world flight without a single real training image? We propose a learning method that we call CAD$^2$RL, which can be used to perform collision-free indoor flight in the real world while being trained entirely on 3D CAD models. Our method uses single RGB images from a monocular camera, without needing to explicitly reconstruct the 3D geometry of the environment or perform explicit motion planning. Our learned collision avoidance policy is represented by a deep convolutional neural network that directly processes raw monocular images and outputs velocity commands. This policy is trained entirely on simulated images, with a Monte Carlo policy evaluation algorithm that directly optimizes the network's ability to produce collision-free flight. By highly randomizing the rendering settings for our simulated training set, we show that we can train a policy that generalizes to the real world, without requiring the simulator to be particularly realistic or high-fidelity. We evaluate our method by flying a real quadrotor through indoor environments, and further evaluate the design choices in our simulator through a series of ablation studies on depth prediction. For supplementary video see: 
\href{https://youtu.be/nXBWmzFrj5s}{https://youtu.be/nXBWmzFrj5s}

\end{abstract}

\IEEEpeerreviewmaketitle

\section{Introduction}
\label{intro}
Indoor navigation and collision avoidance is one of the basic requirements for robotic systems that must operate in unstructured open-world environments, including quadrotors, mobile manipulators, and other mobile robots.
Many of the
most successful approaches to indoor navigation have used 
\vspace{1cm}

{\todo .}

\vspace{5.7cm}
mapping and localization techniques based on 3D perception,  
\noindent including SLAM~\cite{Bachrach09autonomousflight}, depth sensors~\cite{zhang2012microsoft}, stereo cameras~\cite{stereo}, and monocular cameras using structure from motion~\cite{monocular}. The use of sophisticated sensors and specially mounting multiple cameras on the robot imposes additional costs on a robotic platform, which is a particularly prominent issue for weight and power constrained systems such as lightweight aerial vehicles. Monocular cameras, on the other hand, require 3D estimation from motion, which remains a challenging open problem despite considerable recent progress~\cite{engel2014lsd,klein2007parallel}. In this paper, we explore a learning-based approach for indoor navigation, which directly predicts collision-free motor commands from monocular images, without attempting to explicitly model or represent the 3D structure of the environment. In contrast to previous learning-based navigation work~\cite{bills2011autonomous}, our method uses reinforcement learning to obtain supervision that accurately reflects the actual probabilities of collision, instead of separating out obstacle detection and control. The probability of future collision is predicted from raw monocular images using deep convolutional neural networks.

Using reinforcement learning (RL) to learn collision avoidance, especially with high-dimensional representations such as deep neural networks, presents a number of major challenges. First, RL tends to be data-intensive, making it difficult to use with platforms such as aerial vehicles, which have limited flight time and require time-consuming battery changes. Second, RL relies on trial-and-error, which means that, in order to learn to avoid collisions, the vehicle must experience at least a limited number of collisions during training. This can be extremely problematic for fragile robots such as quadrotors.

A promising avenue for addressing these challenges is to train policies in simulation, but it remains an open question whether simulated training of vision-based policies can generalize effectively to the real world. In this work, we show that we can transfer indoor obstacle avoidance policies based on monocular RGB images from simulation to the real world by using a randomized renderer, without relying on an extremely high degree of realism or visual fidelity. Our renderer forces the network to handle a variety of obstacle appearances and lighting conditions, which makes the learned representations invariant to surface appearance. As the result, the network learns geometric features and can robustly detect open spaces.

In contrast to prior work on domain adaptation \cite{rusu2016sim,tzeng2015towards},
our method does not require any real images during training. We demonstrate that our approach can enable navigation of real-world hallways by a real quadrotor using only a monocular camera, without depth or stereo. By training entirely in simulation, we can also use a simple and stable RL algorithm that exploits the ability to reset the environment to any state. Figure~\ref{fig:teaser} shows a diagram of our CAD$^2$RL algorithm.
The algorithm evaluates multiple actions at each state using the current policy, producing dense supervision for the Q-values at that state. Training the Q-function to regress onto these Q-values then corresponds to simple supervised learning. This algorithm sidesteps many of the hyperparameter tuning challenges associated with conventional online RL methods, and is easy to parallelize for efficient simulated training.

The main contribution of our work is an approach for training collision avoidance policies for indoor flight using randomized synthetic environments and deep RL. We designed a set of synthetic 3D hallways that can be used to generate large datasets of randomized scenes, with variable furniture placement, lighting, and textures. Our synthetic data is designed for the task of indoor robot navigation and can also be used as a testbed for RL algorithms. Our proposed RL method is also a novel contribution of this work, and is particularly simple and well-suited for simulated training.

We present an extensive empirical evaluation that assesses generalization to the real world, as well as ablations on a supervised proxy task that studies which aspects of the randomized simulation are most important for generalization.
Our simulated comparative evaluation shows that our approach outperforms several baselines, as well as a prior learning-based method that predicts turning directions~\cite{giusti2016machine}. Our real-world experiments demonstrate the potential for purely simulation-based training of deep neural network navigation policies. Although the policies trained entirely in simulation do experience some collisions in the real world, they outperform baseline methods and are able to navigate effectively around many kinds of obstacles, using only monocular images as input. We therefore conclude that simulated training is a promising direction for learning real-world navigation for aerial vehicles as well as other types of mobile robots.

\section{Related Work}
\label{related}
Any robotic system that must traverse indoor environments is required to perform basic collision avoidance. Standard methods for collision-free indoor navigation take a two step approach to the problem: first map out the local environment and determine its geometry, and then compute a collision-free path for reaching the destination \cite{handbook_of_robotics}. This approach benefits from independent developments in mapping and localization as well as motion planning~\cite{smmk-vbsea-13,mkd-vbcqp-14,barry2015pushbroom}. The 3D geometry of the local environment can be deduced using SLAM with range sensors \cite{Bachrach09autonomousflight}, consumer depth sensors \cite{zhang2012microsoft,hkhrf-rgbdm-12}, stereo camera pairs \cite{stereo}, as well as monocular cameras \cite{monocular}. In~\cite{michels2005high}, laser range scanned real images are used to estimate depth in a supervised learning approach and then the output is used to learn control policies. In~\cite{gupta2017cognitive} simultanouse mapping and planning using RGB-D images is done via a memory network. Reconstruction from monocular images is particularly challenging, and despite considerable progress~\cite{klein2007parallel,engel2014lsd}, remains a difficult open problem. In a recent approach, IM2CAD, CAD model of a room is generated from a single RGB image~\cite{izadinia2017im2cad}. While the synthetic data generated by~\cite{izadinia2017im2cad} could be used for various robotics simulations, the computational overhead makes it less suitable for autonomous indoor flight, where quick inference for finding open spaces is more critical than categorical exact 3D models. 

In our work, we sidestep the challenges of 3D reconstruction by proposing a learning algorithm that can directly predict the probability of collision, without an explicit mapping phase. Learning has previously been used to detect obstacles for indoor flight \cite{bills2011autonomous,kim2015deep}, as well as to directly learn a turn classifier for outdoor forest trail following \cite{giusti2016machine}. In contrast to the work of \cite{bills2011autonomous}, our method directly learns to predict the probability of collision, given an image and a candidate action, without attempting to explicitly detect obstacles. However, our approach still affords considerable flexibility in choosing the action: a higher-level decision making system can choose any collision-free action based, for example, on a higher-level navigational goal. This is in contrast to the prior work, which simply predicts the action that will cause the vehicle to follow a trail \cite{giusti2016machine}. Unlike \cite{giusti2016machine}, our method does not require any human demonstrations or teleoperation.

Besides presenting a deep RL approach for collision avoidance, we describe how this method can be used to learn a generalizable collision predictor in simulation, such that it can then generalize to the real world. Simulated training has been addressed independently in the computer vision and robotics communities in recent years. In computer vision, a number of domain adaptation methods have been proposed that aim to generalize perception systems trained in a source domain into a target domain \cite{tzeng2015simultaneous,NIPS2014_5418}. In robotics, simulation to real-world generalization has been addressed using hierarchies of multi-fidelity simulators \cite{cutler2014reinforcement}, priors imposed on Bayesian dynamics models \cite{cutler2015efficient}. At the intersection of robotics and computer vision, several works have recently applied domain adaptation techniques to perform transfer for robotic perception systems \cite{tzeng2015towards,rusu2016sim,ross2013learning}. In contrast to these works, our method does not use any \emph{explicit} domain adaptation. Instead, we show how the source domain itself can be suitably randomized in order to train a more generalizable model, which we experimentally show can make effective predictions on a range of systematically different target domains.

Our method combines deep neural networks for processing raw camera images \cite{lecun1989backpropagation} with RL. 
In the seminal work of \citet{pomerleau1989alvinn}, a fully connected neural network is used for generating steering commands for the task of road following using raw pixels and laser range finder. Recently, a similar approach was proposed by \cite{bojarski2016end} for a self-driving car. We also generate direction commands from raw visual inputs. However, unlike these prior works, we use RL and do not require any human demonstration data. Furthermore, our method commands the vehicle in 3D, allowing it to change both heading and altitude. Vision-based RL has previously been explored in the context of Q-iteration \cite{m-nfq-05}, and more recently for online Q-learning using temporal-difference algorithms \cite{mnih2015human}. However, these methods were evaluated primarily on synthetic video game domains. Several recent works have extended deep RL methods to real-world robotics applications using either low-dimensional estimated state \cite{chen2016decentralized}  or by collecting an exhaustive real-world dataset under gridworld-like assumptions \cite{zhang2016deeprl}.
In contrast, we propose a simple and stable deep RL algorithm that learns a policy from raw monocular images and does not require seeing any images of the real-world test environment.

\section{Collision Avoidance via Deep RL}
\label{approach}
Our aim is to choose actions for indoor navigation that avoid collisions with obstacles, such as walls and furniture. While we do not explicitly consider the overall navigation objective (e.g. the direction that the vehicle should fly to reach a goal), we present a general and flexible collision avoidance method that predicts which actions are more or less likely to result in collisions, which is straightforward to combine with higher-level navigational objectives. The input to our model consists only of monocular RGB images, without depth or other sensors, making it suitable for low-cost, low-power platforms, though additional sensory inputs could be added in future work. Formally, let $\ot$ denote the camera observation at time $t$, and let $\at$ denote the action, which we will define in Section~\ref{perecption_control}. The goal of the model is to predict the Q-function $Q(\ot, \at)$:
\begin{equation}
\label{returnvalue}
Q(\ot,\at) = \sum_{s=t,\act\sim\pi}^{t+H} {\disc}^{s-t} \mathcal{R}(\os, \as),
\end{equation}
where $\disc \in (0,1)$ is the discount factor, and actions are assumed to be chosen by the current policy $\pi$, which we discuss in Section~\ref{perecption_control}. The horizon $H$ should ideally be $\infty$, but in practice is chosen such that $\disc^H$ is small. $\mathcal{R}$ is the reward function and is equal to zero if collision event happens. Collisions are assumed to end the episode, and therefore can occur only once. Otherwise, the reward at time $s$ is defined as $\min(1,\frac{d_s - r}{\tau_d - r})$, where $r$ is the radius of the vehicle, $d_s$ is the distance to the nearest obstacle at time $s$, and $\tau_d$ is a small threshold distance. This reward function encourages the vehicle to stay far from any obstacles. We could also use the latest Q-function estimate to label the last time step $t+H$, but we found this to be unnecessary to obtain good results. $Q(\ot, \at)$ is learned using reinforcement learning, from the agent's own experience of navigating and avoiding collisions. Once learned, the model can be used to choose collision-free actions $\at$ simply by maximizing the Q-function. 
Training is performed entirely in simulation, where we can easily obtain distances to obstacles and simulate multiple different actions to determine the best one. By randomizing the simulated environment, we can train a model that generalizes effectively to domains with systematic discrepancies from our training environment. We will first describe the formulation of our model and reinforcement learning algorithm, and then present details of our simulated training environment.

\subsection{Perception-Based Control}
\label{perecption_control}

Our perception-based policy uses an action representation that corresponds to positions in image space. The image $\ot$ is discretized into an $M \times M$ grid of bins, and each bin has a corresponding action, such that $\at$ is simply the choice of bin. Once chosen, the bin is transformed into a velocity command $\vel_t$, which corresponds to a vector from the camera location through the image plane at the center of the bin $\at$, normalized to a constant target speed. Intuitively, choosing a bin $\at$ causes the vehicle to fly in the direction of this bin in image space. A greedy policy can use the model $Q(\ot, \at)$ to choose the action with the highest expected reward. We will use $\pi(\obs) = \act$ to denote this policy.

This representation provides the vehicle with enough freedom to choose any desired navigation direction, ascend and descent to avoid obstacles, and navigate tight turns. One advantage of this image-space grid action representation is the flexibility that it provides for general navigational objectives, since we could easily choose the bin using a higher-level navigational controller, subject to the constraint that the probability of collision not exceed some user-chosen threshold. However, in order to evaluate the method in our experiments, we simply follow the greedy strategy.

\subsection{Initialization via Free Space Detection}
\label{freespace_init}
In order to initialize our model with a reasonable starting policy, we use a heuristic pre-training phase based on free space detection. In this pretraining phase, the model is trained to predict $P(l | \ot, \at)$, where $l \in \{0,1\}$ is a label that indicates whether a collision detection raycast in the direction $\vel_t$ corresponding to $\at$ intersects an obstacle. The raycast has a fixed length of $1$ meter. This is essentially equivalent to thresholding the depth map by one meter. This initialization phase roughly corresponds to the assumption that the vehicle will maintain a predefined constant velocity $\vel_t$ . The model, which is represented by a fully convolutional neural network as described in Section~\ref{sec:network}, is trained to label each bin with the collision label $l$, analogously to recent work in image segmentation \cite{chen14semantic}.
The labels are obtained from our simulation engine, as described in Section~\ref{simulation}.

\begin{figure*}[t]
      \centering
      \includegraphics[width=0.97\textwidth]{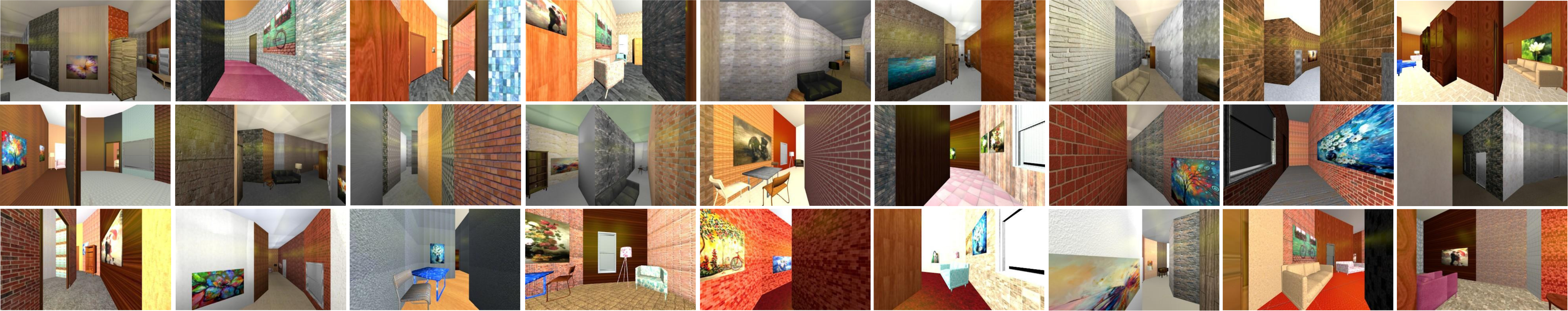}
    	\vspace{-0.11in}
      \caption{Examples of rendered images using our simulator. We randomize textures, lighting and furniture placement to create a visually diverse set of scenes.}
      \label{fig:sample_diversity}
    	\vspace{-0.23in}
   \end{figure*}

\subsection{Reinforcing Collision Avoidance}

\begin{figure}
      \centering
      \includegraphics[width=0.49\textwidth]{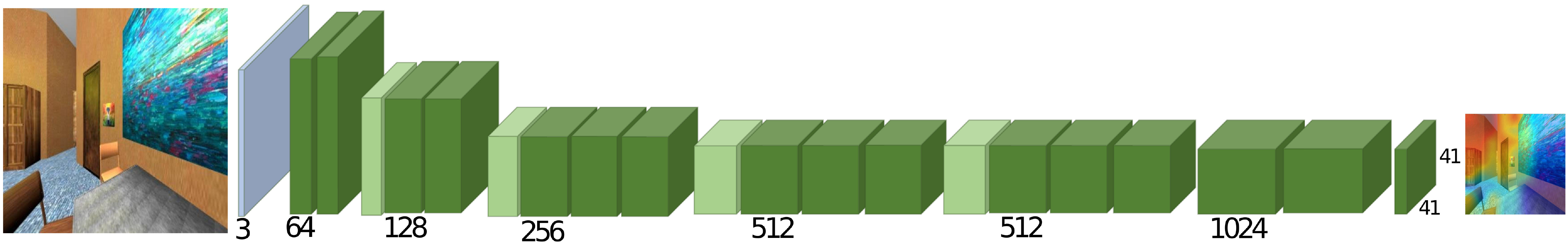}
    	\vspace{-0.28in}
      \caption{We use a fully convolutional neural network to learn the Q-function. Our network, shown above, is based on VGG16 with dilated operations.}
      \label{fig:netarch}
    	\vspace{-0.23in}
   \end{figure}
   
The initial model can estimate free space in front of the vehicle, but this does not necessarily correspond directly to the likelihood of a collision: the vehicle might be able to maneuver out of the way before striking an obstacle within $1$ meter, or it may collide later in the future even if there is sufficient free space at the current time step, for example because of a narrow dead-end. We therefore use deep reinforcement learning to finetune our pretrained model to accurately represent $Q(\ot, \at)$, rather than $P(l | \ot, \at)$. 
To this end, we simulate multiple rollouts by flying through a set of training environments using our latest policy. 
Our score map of $M \times M$ bins, explained in~\ref{perecption_control}, determines the space of actions. Based on our score map, we consider a total of $M^2$ actions $ \act=\{\acti^1,...,\acti^{M^2}\}$ that can be taken after perceiving each observation $\obs$. To generate the training set at each iteration, we sample a collection of states by placing the agent at a random location and with random orientation and generate a rollout of size $K$, given by $(\obs_0,\act_0,\obs_1,\act_1,...,\act_{K-1},\obs_K)$. These states should in principle be obtained from the state distribution of the current policy. Using the model obtained from our pretraining step, the initial policy is simply $\argmax_{i\in\{1,...,M^2\}}{P(l|\obs,\acti^i)}$. We found that we could obtain good performance by sampling the states independently at random, though simply running the latest policy starting from an initial state distribution would also be a simple way to obtain the training states. Once the training states are obtained, we perform
$M \times M$ rollouts from \emph{each} training state using the policy $\pi$ for every possible action $\acti^i,i\in\{1,...,M^2\} $ and evaluate the return of $\acti^i$ according to Equation~(\ref{returnvalue}). 
Since evaluating Equation~(\ref{returnvalue}) requires rolling out the policy for $H$ steps for every action, we choose $H=5$ to reduce computation costs, and instead use a simple approximation to provide smooth target values for $Q(\obs,\acti^i)$. 

This policy evaluation phase provides us with a dataset of observation, action, and return tuples $(\obs_t, \at, Q(\ot,\at))$, which we can use to update the policy. Since we evaluate every action for each image $\ot$, the dataset consists of densely labeled images with $Q$ values reflecting the expected sum of future rewards for the current policy $\pi$.

Our method can be interpreted as a modification of fitted Q-iteration~\cite{m-nfq-05}, in the sense that we iteratively refit a Q-function estimator to samples, as well as a variant of modified policy iteration (MPI)~\cite{puterman1978modified} or Monte Carlo policy evaluation, in the sense that we estimate Q-values using multi-step rollouts of the current policy. 
To our knowledge, ours is the first algorithm of this class to be extended to deep reinforcement learning with raw image inputs. The particular details of the approach, including the evaluation of each action at each state, are specifically designed for our simulated training setup to exploit the capabilities of the simulation and provide for a simple and stable learning algorithm. We perform rollouts in simulated training hallways. This allows us to perform multiple rollouts from the state at each time step, perform ground truth collision detection raycasts for pretraining, and removes concerns about training-time collisions. Unlike conventional RL methods that perform rollouts directly in the test environment~\cite{mnih2015human}, we perform rollouts in simulated training hallways. However, this also means that our model must have generalization from the simulated training hallways to real-world environments at test time. To that, we developed a randomized simulated training environment, which we describe in the next section.

\subsection{Network Architecture}
\label{sec:network}

In order to represent the Q-function and the initial open space predictor, we use a deep fully convolutional neural network with dilation operations, built on the VGG16~\cite{VGG16} architecture following~\cite{chen14semantic} as shown in Figure~\ref{fig:netarch}. The output score map corresponds to a grid of $41\times 41$ bins, which constitutes the action space for deep reinforcement learning. The network is trained with stochastic gradient descent (SGD), with a cross-entropy loss function.

\section{Learning from Simulation}
\label{simulation}
Conventionally, learning-based approaches to autonomous flight have relied on learning from demonstration~\cite{helicopter-IJRR2010,acqn-arlah-06,pa-dlhdm-15,ross2013learning}.
Although the learning by demonstration approach has been successfully applied to a number of flight scenarios, the requirement for human-provided demonstrations limits the quantity and diversity of data that can be used for training. Since dataset size has been demonstrated to be critical for the success of learning methods, this likely severely limits the generalization capacity of purely demonstration-based methods. If we can train flight controllers using larger and more diverse datasets collected autonomously, we can in principle achieve substantially better generalization. 
However, in order to autonomously learn effective collision prediction models, the vehicle needs to see enough examples of collisions during training to build an accurate estimator. This is problematic in real physical environments, where even a single collision can lead to damage or loss of the vehicle. To get the benefits of an autonomous learning from the agent's own experience and overcome the limitations of data collection in learning from demonstration method, we use a simulated training environment that is specifically designed to enable effective transfer to real-world settings.
\begin{figure}
      \centering
      \includegraphics[width=0.48\textwidth]{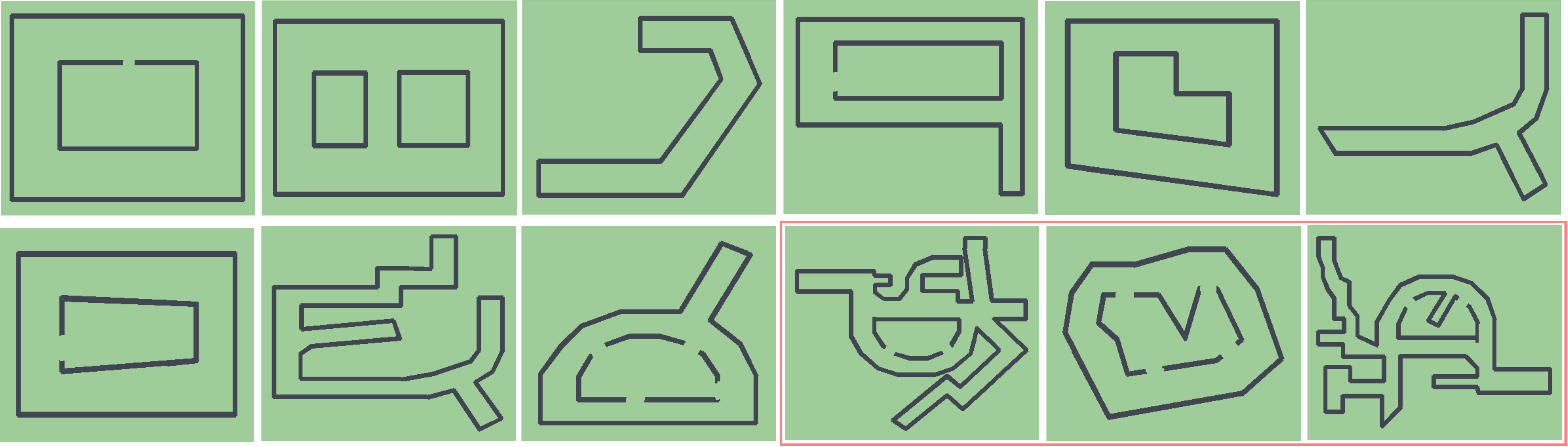}
      \vspace{-0.1in}
      \caption{Floor plans of the synthetic hallways. The last three hallways are used for evaluation while the first 9 are used during training.}
      \label{fig:training_halls}
      \vspace{-0.23in}
   \end{figure}

We manually designed a collection of 3D indoor environments to form the basis of our simulated training setup. The environments were built using the Blender~\cite{blender} open-source 3D modeling suite.
Our synthetic dataset contains different hallways, shown in Figure~\ref{fig:training_halls}, and represent a variety of structures that can be seen in real hallways, such as long straight or circular segments with multiple junction connectivity, as well as side rooms with open or closed doors. We use furnitures with various type and size to populate the hallways. The walls are textured with randomly chosen textures(e.g. wood, metal, textile, carpet, stone, glass, etc.), and illuminated with lights that are placed and oriented at random. In order to provide a diversity of viewpoints we render pretraining images by flying a simulated camera with randomized height and random camera orientation.

The randomization of the hallway parameters produces a very large diversity of training scenes, a sample of which can be seen in Figure~\ref{fig:sample_diversity}. Although the training hallways are far from being photo-realistic, the large variety of appearances allows us to train highly generalizable models, as we will discuss in the experimental evaluation. The intuition behind this idea is that, by forcing the model to handle a greater degree of variation than is typical in real hallways (e.g., wide ranges of lighting conditions and textures, some of which are realistic, and some not), we can produce a model that generalizes also to real-world scenes, which might be systematically different from our renderings. That is, the wider we vary the parameters in simulation, the more likely we are to capture properties of the real world somewhere in the set of all possible scenes we consider. Our findings in this regard are aligned with the results obtained in other recent works~\cite{richter2016playing}, which also used only synthetic renderings to train visual models, but did not explicitly consider wide-ranging randomization of the training scenes.

\section{Experimental Results}
\label{experiment}
Despite that reinforcement learning evaluations emphasize mastery over generalization here our focus is on to evaluate the generalization capability of our proposed approach. Testing generalization is specially important from robotics perspective since the autonomous agent should be able to generalize to the diverse real-world settings. To this end, we evaluate our performance by running several experiments both in synthetic and real environments none of which had been seen during the training time.
We compared our results against a set of baselines and also qualitatively evaluate our performance in various real-world scenarios. Additionally, we present an ablation study on a real-world RGB-D dataset to quantitatively evaluate our proposed randomized simulator for simulation to real-world transfer.
In all the experiments (synthetic and real-world flights), CAD$^2$RL is trained on a fixed set of synthetic 3D models of hallways and in a fully simulated environment without being exposed to any real images.

\subsection{Realistic Environment Evaluation}
\label{sec:real_exp}  
\begin{figure}[t]
    \centering
    \includegraphics[width=0.45\textwidth,trim={0.5cm 0.2cm 0.95cm 1.07cm},clip]{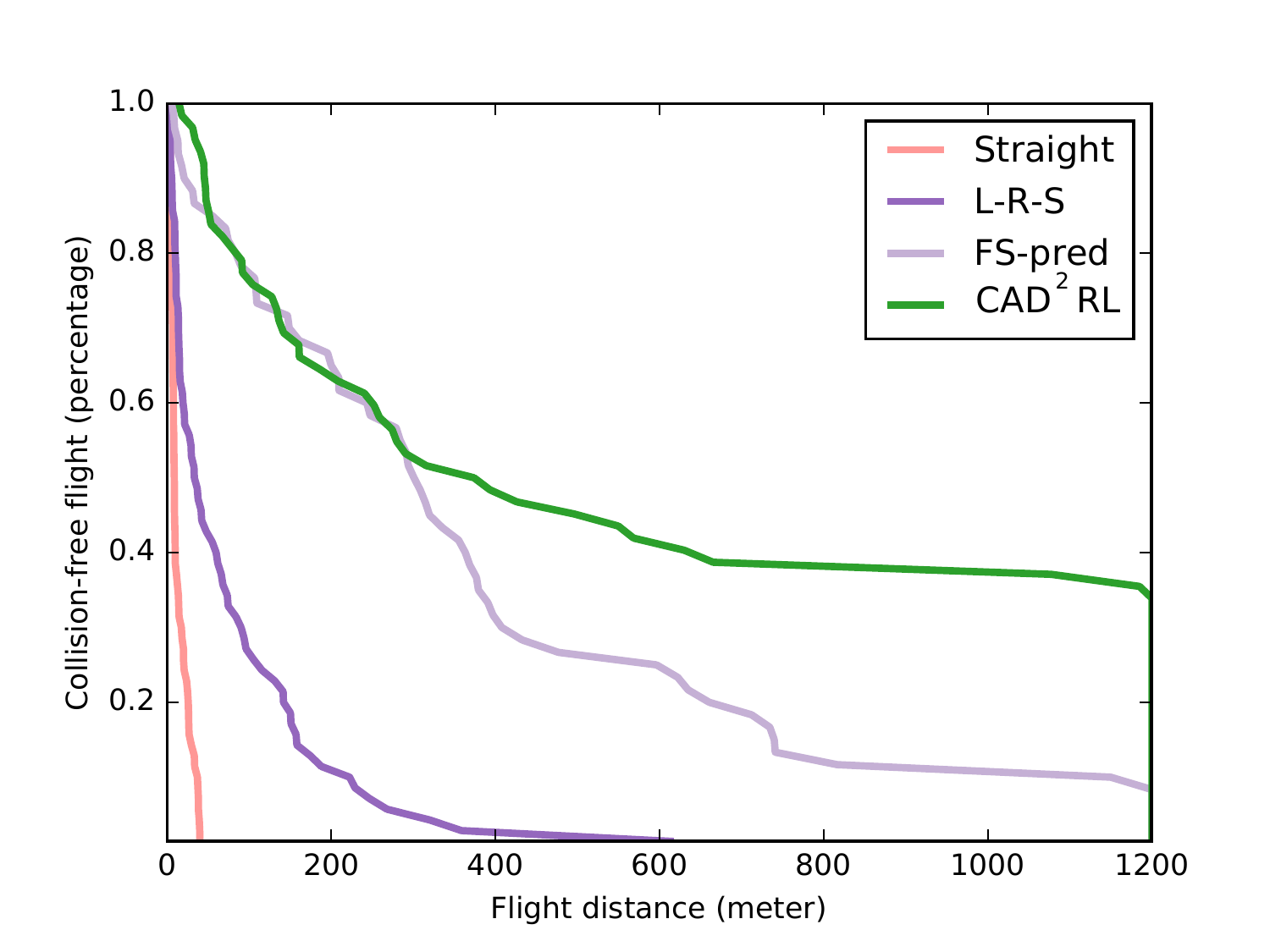}
    	\vspace{-0.15in}
    \caption{Quantitative results on a realistically textured hallway. Our approach, CAD$^2$RL, outperforms the prior method (L-R-S) and other baselines.}
    \label{real_quantitative}
    \vspace{-0.25in}
\end{figure}

In order to evaluate how well such a model might transfer to a realistic environment, we used a realistic 3D mesh provided by~\cite{kua2012automatic}. Testing on this data can provide us a close proxy of our performance in a real indoor environment and also evaluates the generalization capability of our method in a systematically different environment than our training environments. Figure~\ref{fig:real_maze_qualitative} shows the floorplan of this hallway, as well as several samples of its interior view. 
We generated $60$ random initialization point from various locations in the hallways. These points are fixed and all baselines are evaluated on the same set of points so that their performance is directly comparable. Figure~\ref{fig:real_maze_qualitative}.a depicts the initialization points as red dots. The velocity of the quadrotor is fixed to $0.2$ meters per time step in this experiment, and the maximum number of steps is set to $6000$ which is equal to $1.2$ kilometers.

Our aim is to evaluate the performance of our trained policy in terms of the duration of collision free flight. To do this, we run continuous episodes that terminate upon experiencing a collision, and count how many steps are taken before a collision takes place. We set the maximum number of steps to a fixed number throughout each experiment. We evaluate performance in terms of the percentage of trials that reached a particular flight length. To that end, we report the results using a curve that plots the distance traveled along the horizontal axis, and the percentage of trials that reached that distance before a collision on the vertical axis. This provides an accurate and rigorous evaluation of each policy, and allows us to interpret for each method whether it is prone to collide early in the flight, or can maintain collision-free flight at length. Note that achieving completely collision-free flight in all cases from completely randomized initial configurations is exceptionally difficult.

In this experiment, we compare against two baselines explained below. We also report the performance of our base Free Space prediction (FS-pred) controller to analyze the improvement obtained by incorporating deep reinforcement learning. In the FS-pred, the model described in~\ref{freespace_init} is used.  

\noindent{\bf Straight Controller} This lower bound baseline flies in a straight line without turning. In a long straight hallway, this baseline establishes how far the vehicle can fly without any perception, allowing us to ascertain the difficulty of the initialization conditions.

\noindent{\bf Left, Right, and Straight (LRS) Controller}
This baseline, based on~\cite{giusti2016machine}, directly predicts the flight direction from images. The commands are discretized into three bins: ``left,'' ``right,'' or ``straight,'' and the predictions are made by a deep convolutional neural network from raw images. For training the model, prior work used real-world images collected from three cameras pointing left, right and straight that were carried manually through forest trails. We simulated the same training setup in our training environments. We finetuned a VGG16~\cite{VGG16} model, pretrained with ImageNet classification. This method can be considered a human-supervised alternative to our autonomous collision avoidance policy.

\subsubsection{Quantitative Evaluation}

Figure~\ref{real_quantitative} summarizes the performance of our proposed CAD$^2$RL method compared with other baselines. Our method outperforms the prior methods and baselines by a substantial margin. Qualitatively, we found that the LRS method tends to make poor decisions at intersections, and the coarse granularity of its action representation also makes it difficult for it to maneuver near obstacles. CAD$^2$RL is able to maintain a collision-free flight of $1.2$ kilometers in about $40\%$ of the cases, and substantially outperforms the model that is simply trained with supervised learning to predict $1$ meter of free space in front of the vehicle. This experiment shows that although we did not use real images during training, our learned model can generalize to substantially different and more realistic environments, and can maintain collision-free flight for relatively long periods.

\subsubsection{Qualitative Evaluation}
\begin{figure*}
      \centering
      \includegraphics[width=0.96\textwidth,height=0.37\textwidth]{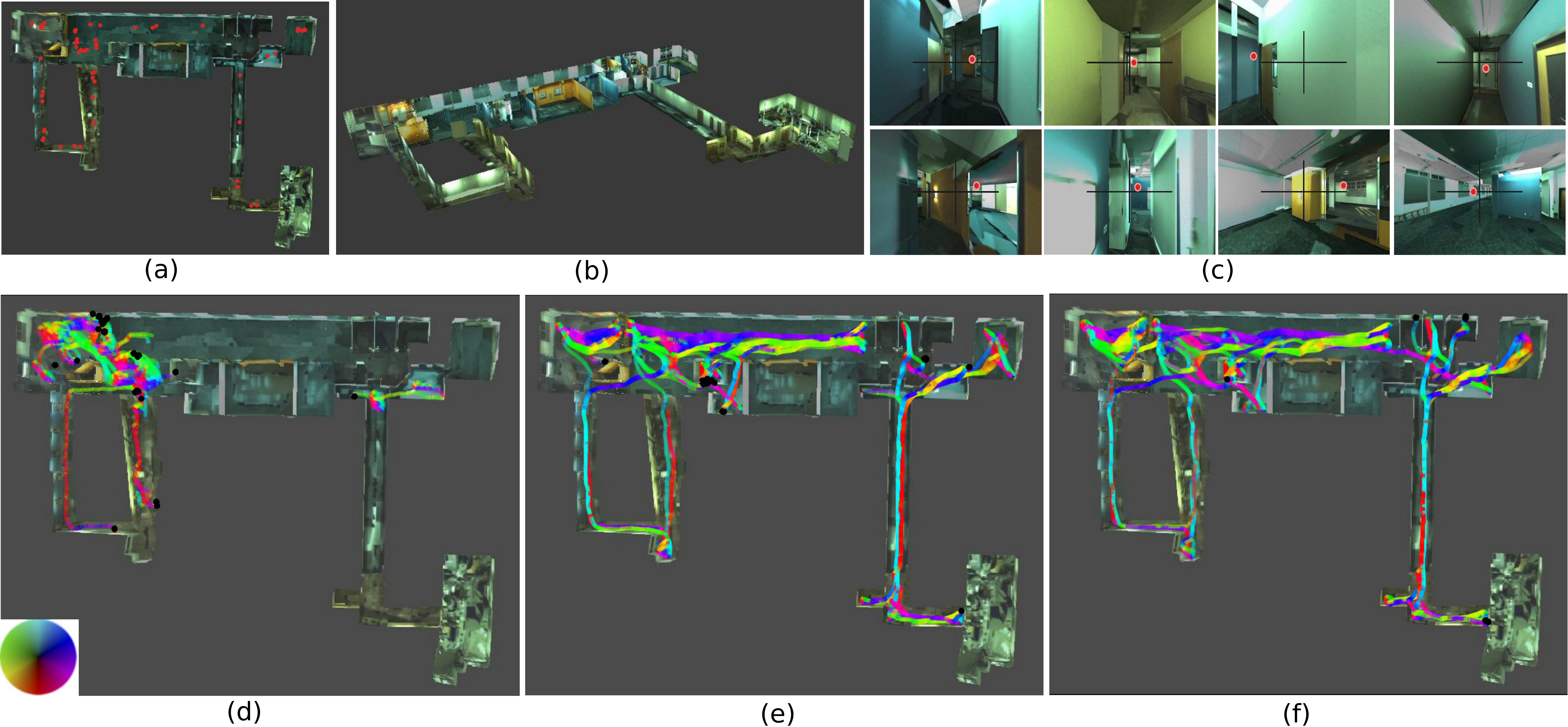}
    	\vspace{-0.15in}
      \caption{Qualitative results on a realistically textured hallway. Colors correspond to the direction of trajectory movement at each point in the hallway as per the color wheel. (a) Red dots show flight initialization points (b) Overlook view of the hallway (c) Red dots show the control points produced by CAD$^2$RL. (d) LRS trajectories (e) Perception controller (FS-pred) trajectories (f) CAD$^2$RL trajectories.}
      \label{fig:real_maze_qualitative}
    \vspace{-0.1in}
\end{figure*}
To be able to qualitatively compare the performance and behavior of CAD$^2$RL with our perception based controller and the LRS method, we visualized the trajectory of the flights overlaid on the floor-plan of the hallway as shown in Figure~\ref{fig:real_maze_qualitative}. For this purpose, we sorted the trajectories of each method based on the traveled distance and selected the top $25$ longest flights from each method. The trajectory colors show the flight direction at each point. 
The black dots indicate the locations of the hallway where collisions occurred. This visualization shows that CAD$^2$RL could maintain a collision-free flight in various locations in the hallway and has fewer collisions at the dead-ends, corners, and junctions compared with the other two methods. LRS often experienced collisions in corners and is more vulnerable to bad initial locations. The policy trained with free space prediction outperformed the LRS method, but often is trapped in rooms or fail near junctions and corners. This illustrates that the controller trained with RL was able to acquire a better strategy for medium-horizon planning, compared to the directly supervised greedy methods.

\subsection{Real World Flight Experiments}

We evaluated our learned collision avoidance model by flying a drone in real world indoor environments. These flights required flying through open spaces, navigating hallways, and taking sharp turns, while avoiding collisions with furniture, walls, and fixtures. 
We used two different drone platforms: the Parrot Bebop 1.0 and the Bebop 2.0, both controlled via the ROS Bebop autonomy package~\cite{bebob}. We perform real-world flight in several different scenarios and evaluate our performance both quantitatively and qualitatively.

\subsubsection{Quantitative Evaluation}
For quantitative evaluation, we ran controlled experiments on the task of hallway following. We fixed all the testing conditions while navigating the drone with either of the CAD$^2$RL and a baseline controller. The testing conditions include the initial velocity, angular speed, drone platform and the test environment.
As was concluded from the experiments in section~\ref{sec:real_exp}, FS-pred was the strongest baseline, and we therefore included it as a comparison in this experiment. 
We ran experiments in two different buildings, Cory Hall and SDH (Sutardja Dai Hall), both located on the UC Berkeley campus. These buildings have considerably different floor plans, wall textures, and lighting conditions, as can be seen in Figure~\ref{real_flight_samples}.c and Figure~\ref{real_flight_samples}.d. 
Our testing environment in Cory Hall contained three turns and two junctions, while the SDH test environment had one turn and one junction. The width of the Cory hall hallway is $\sim3$ meters while the SDH hallway is $\sim2$ meters wide.

Table~\ref{tab:real_exp} summarizes the results. The safe flight time is given by the average length of a collision free flight in terms of distance or time between collisions. CAD$^2$RL experienced fewer collisions and has longer expected safe flight.
This suggests that the CAD$^2$RL policy makes fewer mistakes and is more robust to perturbations and drift. Both methods performed better in Cory, since SDH has narrower hallways with glossy textureless walls as well as stronger air currents. While we fixed the test environment and the flying speed, the traveled distance and time is slightly different from one algorithm to another due to the fact that the algorithms generated different commands and navigated the drone to slightly different locations in the hallways.

\begin{table*}
\begin{small}
\begin{center}
\caption{\small Real world flight results.\vspace{-.05in}}
\renewcommand{\arraystretch}{0.1}
\begin{tabular}[10pt]{lccccccc} 
\toprule
Environment & {Traveled Distance} & {Travel Time}  & {Collision} & {Collision} & {Safe Flight } & {Safe Flight} & {Total}  \\[-.02ex]
 & {(meters)} & {(minutes)}  & {(per meter)} & {(per minute)} & {(meters)} & {(minutes)} & {Collisions}  \\
\midrule
Cory FS-pred& {162.458} & {12.01} & {0.080} & {1.081}& {12.496} & {0.924} & {13}  \\
Cory CAD$^2$RL & {163.779} & {11.950} & {0.0366} & {0.502} & {27.296} & {1.991} & {6}  \\
\midrule
SDH FS-pred& {53.492} & {4.016} & {0.130} & {1.742} & {7.641} & {0.573} & {7}  \\
SDH CAD$^2$RL & {54.813} & {4.183} & {0.072} & {0.956} & {13.703} & {1.045} & {4}  \\
\bottomrule
\end{tabular}
\end{center}
\end{small}
\label{tab:real_exp}
\vspace{-.3in}
\end{table*}

\subsubsection{Qualitative Evaluation}
 \begin{figure*}
     \centering
     \includegraphics[width=0.88\textwidth]{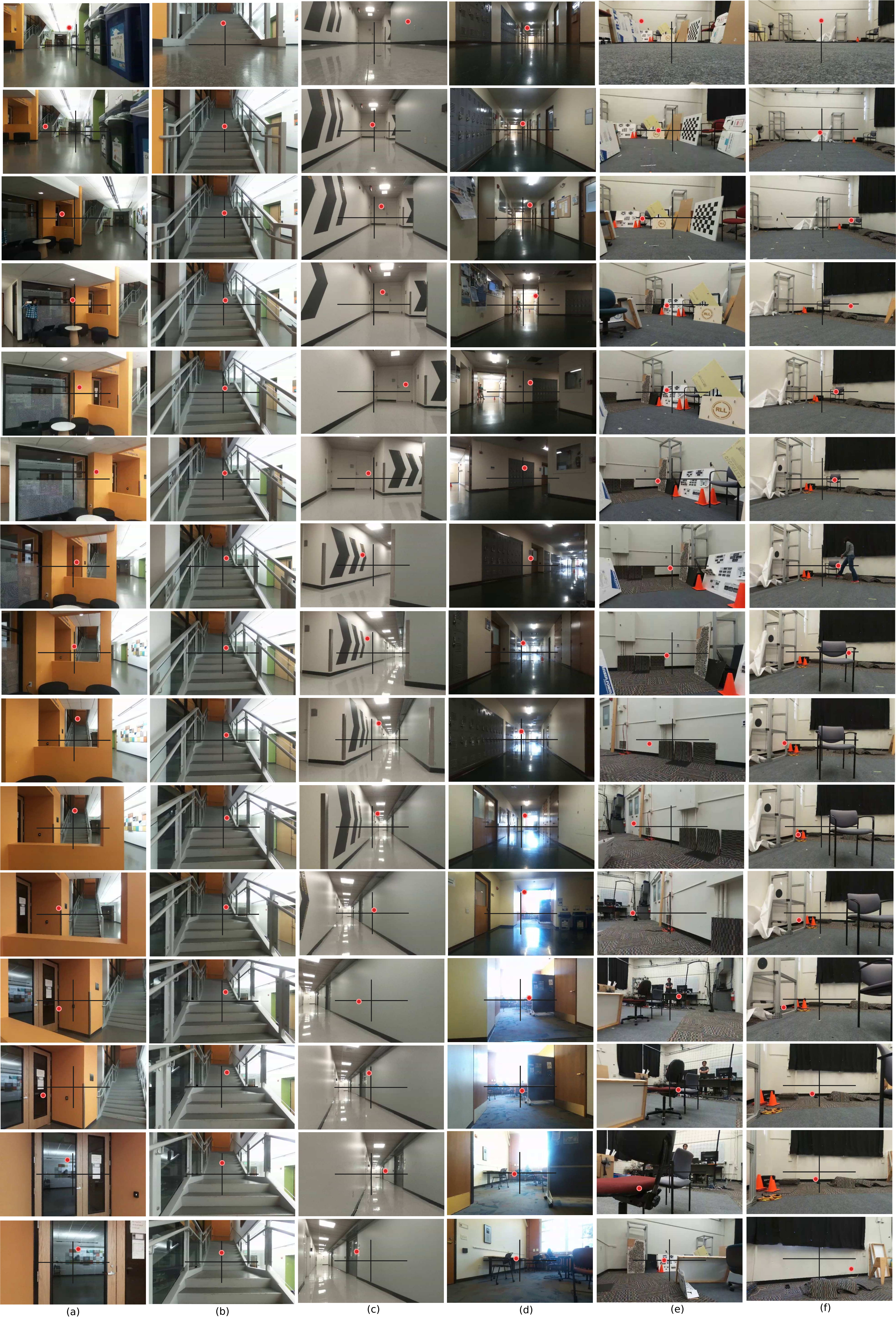}
     \vspace{-0.15in}
     \caption{Snapshots of autonomous flight in various real indoor scenarios. Frames ordered from top to bottom. Red dots show the commanded flight direction by CAD$^2$RL. (a) Flying near furniture, around corners, through a window; (b) Flying up a staircase; (c) Navigating in narrow corridors; (d) Navigating through junctions, fly through rooms; (e) Flying through a maze of random obstacles in a confined space; (f) Avoiding dynamic obstacles.}
     \label{real_flight_samples}
     \vspace{-0.1in}
 \end{figure*}
We performed real world flight in various indoor scenarios. We briefly explain each scenario and sequence snapshots are shown in Figure~\ref{real_flight_samples}. 

\noindent{\bf (a) Flying near furniture, around corners, and through a window:}
As shown in Figure~\ref{real_flight_samples}.a. the drone starts from one end of a hallway connected to a small lounge area with furniture.
 The drone first flies toward the open lounge area, and then turns toward a corner of the room. There, it detects an opening in the wall which is visually similar to an open doorway or window, and adjust its height to fly through it.
The drone then encounters a reflective glass door, which reflects the hallway behind it. Since no such structures were present during training, the reflective door fools the controller, causing the drone to crash into the door. Note that the controller navigates multiple structures that are substantially different, both visually and geometrically, from the ones encountered during simulated training. 

\noindent{\bf (b) Flying up a staircase:} Here, our goal is to evaluate the generalization capability of the controller to changes in elevation. A staircase provides a good example of this. To avoid colliding with the stairs, the drone must continuously increase altitude. As can be seen from the snapshots in the Figure~\ref{real_flight_samples}.b, the controller produces actions that increase the altitude of the drone at each step along the staircase. Since we used an altitude limit for safety reasons, the drone only flew halfway up the staircase, but this experiment shows that the controller could effectively generalize to structures such as staircases that were not present during training.

\noindent{\bf (c) Navigating through narrow corridors:} In this scenario, the drone flies through a corridor. The drone successfully takes a turn at Frames 1-4 in Figuree~\ref{real_flight_samples}.c to avoid flying into a dead end. The corridors in this test scenario are narrow ($\sim2$ meters) and have strong air currents due to air conditioning.
n

\noindent{\bf (d) Flying through junctions and rooms:} Here, the drone navigates through a long hallway with junctions. At the end it enters a doorway which is connected to a study room. The controller successfully navigates the drone through the narrow door and into the room without colliding with the chairs. 

\noindent{\bf (e) Flying through a maze of random obstacles in a confined space:}
We built a small U-shaped maze out of low obstacles in the lab. This maze is built using chairs and pieces of board with various appearance and colors. To prevent the drone from simply flying over the obstacles, we limited the altitude to $3$ feet.
Note that flying the drone at low altitude is challenging, as the air turbulence becomes significant and affects the drone's stability. The cardboard shifts due to air turbulence, and the open area is very narrow ($\sim1$ meter), making this a challenging test environment. The sequence in Figure~\ref{real_flight_samples}.e shows that the controller successfully navigates the drone throughout the maze, making a turn near the red chair and turning back into the maze, without colliding. 

\noindent{\bf (f) Avoiding dynamic obstacles:} In this scenario, the drone begins in the lab with no obstacles and an altitude of around $3$ feet. We then place a chair in the path of the drone, as seen in frames 3-4 of Figure~\ref{real_flight_samples}.f. The controller recovers and avoids an imminent collision with the chair, passing it on the left.

The above qualitative evaluation study shows the generalization capability of our trained model and demonstrates the extent of the maneuvering skills learned by CAD$^2$RL. Although our model is specifically trained for the task of hallway navigation, the limited number of furniture items present in simulation also force the policy to be robust to oddly shaped obstacles, and train it to change altitude to avoid collisions. Navigating through the obstacles in the scenarios (a), (b), (e), and (f) required collision avoidance with general obstacles and other than just walls. We observed that our model could perform reasonably well in these cases, and could often recover from its mistakes, though particularly novel situations proved confusing.

\subsection{Ablation Study for Real World Transfer}
\label{sec:kinect}  

\begin{figure}
    \centering
    \includegraphics[width=0.43\textwidth,trim={0.95cm 0.9cm 0.95cm 1.145cm},clip]{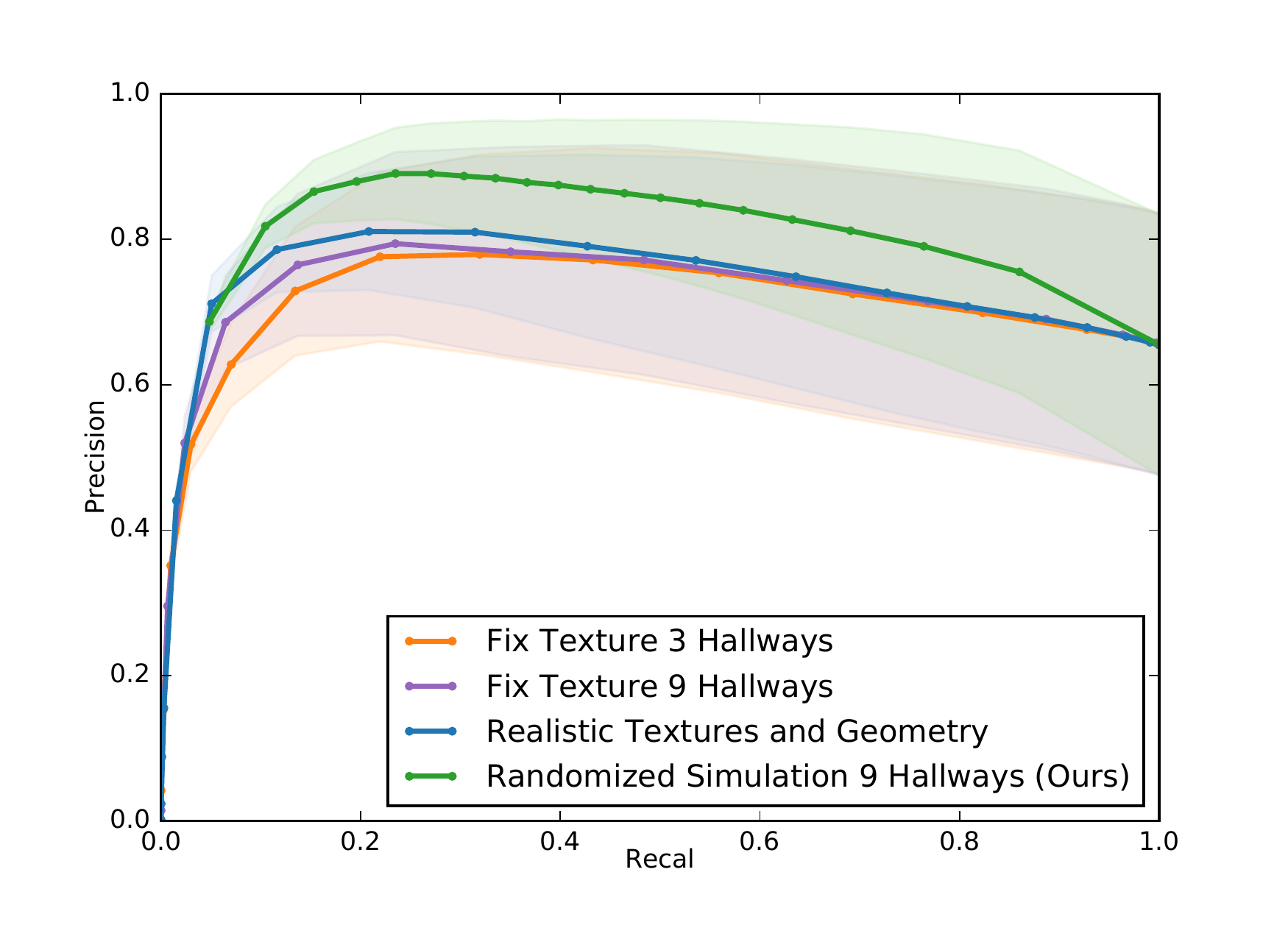}
    	\vspace{-0.15in}
    \caption{Quantitative results for free-space prediction with different simulators. The network trained on randomized hallways outperforms networks trained on less randomized simulations and even on realistic textured hallways.}
    \label{kinect_quantitative}
    \vspace{-0.25in}
\end{figure}

In this section, we present an ablation study to identify how important the randomization of the environment is for effective simulation to real-world transfer. Since conducting statistically significant real-world flight trials for many training conditions is time-consuming and subject to confounding factors (air currents, lighting conditions, etc.), we instead opted for a proxy task that corresponds to free-space prediction from real RGB images, with ground truth labels obtained via a depth camera. The goal in this task is to predict, for each point in the image, whether there is an obstacle within a certain threshold distance of the camera or if the pixel corresponds to free space. Although this proxy task does not directly correspond to collision-free flight, the reduced variance of the evaluation (since all methods are tested on exactly the same images) makes this a good choice for the ablation study. While we obtained reasonably good performance for avoiding collisions in the hallways, more detailed depth estimation ~\cite{saxena2005learning,liu2016learning,liu2015deep} could also be used without loss of generality.

We used the same architecture as in Section~\ref{freespace_init} for the free-space prediction network and trained free-space predictors using rendered images from different simulated setups. We compared the obtained results against a similar network trained using our proposed randomized simulation. We used the same number of images sampled similarly  from various locations in the hallways. The ablated networks are trained with images rendered from (a) a simulator that used {\bf F}ixed {\bf T}extures and lighting, and only 3 of our training hallways (FT3); (b) {\bf F}ixed {\bf T}extures and lighting using all 9 training hallways (FT9) (c) the more {\bf R}ealistic {\bf T}extures and {\bf G}eometry hallway provided by~\cite{kua2012automatic} (RTG) and (d) our approach, with randomized textures and lighting from 9 hallways. While (a), (b), and (d) are captured from synthetic hallways, in (c) the data is captured via a SLAM-based reconstruction system from the Cory Hall in the UC Berkeley campus. Therefore, this data has realistic geometry textured with natural images, and allows us to understand how the method would perform if trained on reconstructed RGBD data.

\begin{figure}[t]
    \centering
    \includegraphics[width=0.46\textwidth,trim={0.95cm 0.95cm 0.95cm .9cm},clip]{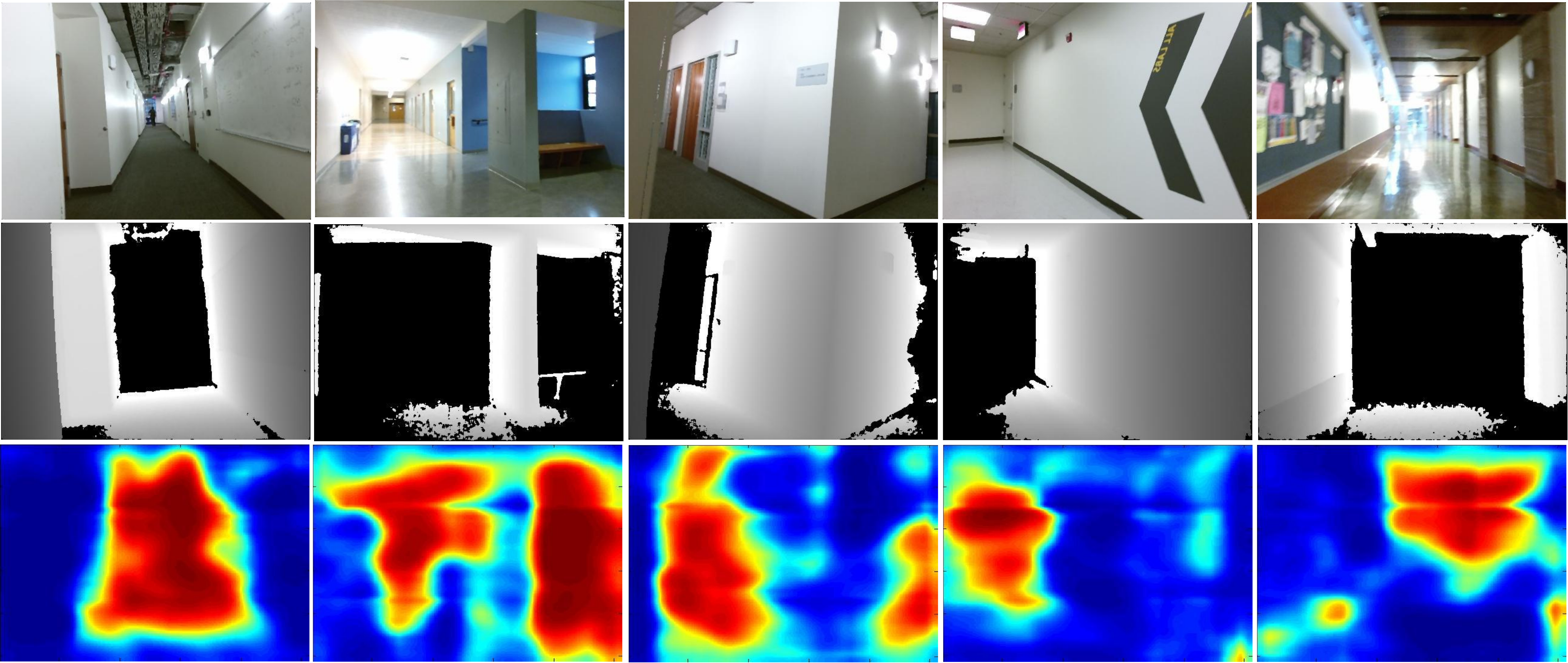}
    	\vspace{-0.13in}
    \caption{Examples of the collected pairs of RGB (top row) and depth (mid row) data for the free-space test set. The free-space probability map predicted by our approach is shown in the bottom row.}
    \label{kinect_data}
    \vspace{-0.28in}
\end{figure}

Our dataset contains RGB-D images captured from 5 hallways, in Cory Hall and SDH (Sutardja Dai Hall) located in UC Berkeley, with various lighting and texture conditions. We used a Kinect v2 and our dataset contains a total of 620 RGB-D images. Several example images of this data are shown in Figure~\ref{kinect_data}.
We used the depth channel to automatically annotate the images with free-space vs. non-free-space labels.

For each pixel in the input image, the network produces a probability value for free-space prediction. To evaluate the accuracy of free-space prediction we sweep a threshold from 0 to 1 to label each pixel using our prediction network. We compute the precision and recall at each threshold and plot the precision-recall curve as the performance metric. Precision is the number pixels correctly labeled as free-space divided by the total number of pixels predicted as free-space, while recall is the the number pixels correctly labeled as free-space divided by the total number pixels belonging to the free-space according to the ground truth. Since we use monocular images, there is a scale ambiguity in the size of hallways as well as in the range of sensible depths, which may not match between the simulated and real images. To overcome this ambiguity and to make a fair comparison, we labeled image pixels (for free-space vs non-free-space) by varying the depth threshold from 1 to 4 meters (steps of $\sim30cm$) and computed the average precision/recall corresponding for each threshold over 13 runs.

Figure~\ref{kinect_quantitative} shows the results, with shaded areas showing the standard deviation in precision. The network trained with the synthetic data rendered by our proposed randomized simulator outperforms the other networks. The images used for FT3 and FT9 are rendered on the same hallways as RT9, except that the textures and lighting are not randomized. As a result, these networks do not learn texture and color invariant features and cannot generalize well to the real images. In RTG, the images are rendered with realistic geometry and textures, and thus they are less affected by the scale ambiguity. Furthermore, the realistic textures in RTG are obtained from similar hallways
as the one we used for our RGB-D test set. Despite this, the network trained on a realistic rendering of the same hallway actually performs worse than the network trained on our randomized simulator, by a substantial margin. For qualitative analyis, we show the probability map of free-space prediction obtained from our approach in the last row of Figure~\ref{kinect_quantitative}. We see that high probabilities are assigned to free spaces. Although the free-space prediction proxy task is not a perfect analogue for collision-free flight, these results suggest that randomization is important for good generalization, and that more realistic renderings should not necessarily be preferred to ones that are less realistic but more diverse.

 \section{Discussion}
 \label{conclusion}
 We presented a method for training deep neural network policies for obstacle avoidance and hallway following, using only simulated monocular RGB images. We described a new simple and stable deep reinforcement learning algorithm for learning in simulation. We also demonstrate that training on randomized simulated scenes produces a model that can successfully fly and avoid obstacles in the real world, and quantitatively evaluated our randomized scenes on a proxy free-space prediction task to show the importance of randomization for real-world transfer. Our simulated evaluation further shows that our method outperforms several baselines, as well as a prior end-to-end learning-based method. Our aim in this work is to evaluate the potential of policies trained \emph{entirely} in simulation to transfer to the real world, so as to understand the benefits and limitations of simulated training. To attain the best results in real environments, future work could combine simulated training with real data. Extending our approach via finetuning or domain adaptation is therefore a promising direction for future work that is likely to improve performance substantially, and lead to effective learned real-world visual navigation policies using only modest amounts of real-world training. Our approach could incorporate data from other sensors, such as depth cameras, which should improve the performance of the learned policies.

 \section*{ACKNOWLEDGMENT}
 \label{ack}
 The authors would like to thank Larry Zitnick for helpful discussions and insightful remarks. This work was made possible by an ONR Young Investigator Program Award and support from Google, NVIDIA, and Berkeley DeepDrive.

{\small
\bibliographystyle{plainnat}
\bibliography{main}

\begin{thebibliography}{44}
\providecommand{\natexlab}[1]{#1}
\providecommand{\url}[1]{\texttt{#1}}
\expandafter\ifx\csname urlstyle\endcsname\relax
  \providecommand{\doi}[1]{doi: #1}\else
  \providecommand{\doi}{doi: \begingroup \urlstyle{rm}\Url}\fi

\bibitem[Abbeel et~al.(2006)Abbeel, Coates, Quigley, and Ng]{acqn-arlah-06}
P.~Abbeel, A.~Coates, M.~Quigley, and A.~Ng.
\newblock An application of reinforcement learning to aerobatic helicopter
  flight.
\newblock In \emph{NIPS}, 2006.

\bibitem[Abbeel et~al.(2010)Abbeel, Coates, and Ng]{helicopter-IJRR2010}
Pieter Abbeel, Adam Coates, and Andrew~Y. Ng.
\newblock Autonomous helicopter aerobatics through apprenticeship learning.
\newblock \emph{IJRR}, 2010.

\bibitem[Bachrach et~al.(2009)Bachrach, He, and
  Roy]{Bachrach09autonomousflight}
Abraham Bachrach, Ruijie He, and Nicholas Roy.
\newblock Autonomous flight in unstructured and unknown indoor environments.
\newblock In \emph{EMAV}, 2009.

\bibitem[Barry and Tedrake(2015)]{barry2015pushbroom}
Andrew~J Barry and Russ Tedrake.
\newblock Pushbroom stereo for high-speed navigation in cluttered environments.
\newblock In \emph{ICRA}. IEEE, 2015.

\bibitem[Bills et~al.(2011)Bills, Chen, and Saxena]{bills2011autonomous}
Cooper Bills, Joyce Chen, and Ashutosh Saxena.
\newblock Autonomous mav flight in indoor environments using single image
  perspective cues.
\newblock In \emph{ICRA}, 2011.

\bibitem[{Blender Community}()]{blender}
{Blender Community}.
\newblock {Blender: Open Source 3D modeling suit}.
\newblock \url{http://www.blender.org}.

\bibitem[Bojarski et~al.(2016)Bojarski, Del~Testa, Dworakowski, Firner, Flepp,
  Goyal, Jackel, Monfort, Muller, Zhang, et~al.]{bojarski2016end}
Mariusz Bojarski, Davide Del~Testa, Daniel Dworakowski, Bernhard Firner, Beat
  Flepp, Prasoon Goyal, Lawrence~D Jackel, Mathew Monfort, Urs Muller, Jiakai
  Zhang, et~al.
\newblock End to end learning for self-driving cars.
\newblock \emph{arXiv preprint arXiv:1604.07316}, 2016.

\bibitem[Celik et~al.(2009)Celik, Chung, Clausman, and Somani]{monocular}
K.~Celik, S.J. Chung, M.~Clausman, and A.~Somani.
\newblock Monocular vision {SLAM} for indoor aerial vehicles.
\newblock In \emph{IROS}, 2009.

\bibitem[Chen et~al.(2015)Chen, Papandreou, Kokkinos, Murphy, and
  Yuille]{chen14semantic}
Liang-Chieh Chen, George Papandreou, Iasonas Kokkinos, Kevin Murphy, and Alan~L
  Yuille.
\newblock Semantic image segmentation with deep convolutional nets and fully
  connected crfs.
\newblock In \emph{ICLR}, 2015.

\bibitem[Chen et~al.(2016)Chen, Liu, Everett, and How]{chen2016decentralized}
Yu~Fan Chen, Miao Liu, Michael Everett, and Jonathan How.
\newblock Decentralized non-communicating multiagent collision avoidance with
  deep reinforcement learning.
\newblock \emph{arXiv preprint arXiv:1609.07845}, 2016.

\bibitem[Cutler and How(2015)]{cutler2015efficient}
Mark Cutler and Jonathan~P How.
\newblock Efficient reinforcement learning for robots using informative
  simulated priors.
\newblock In \emph{ICRA}, 2015.

\bibitem[Cutler et~al.(2014)Cutler, Walsh, and How]{cutler2014reinforcement}
Mark Cutler, Thomas~J Walsh, and Jonathan~P How.
\newblock Reinforcement learning with multi-fidelity simulators.
\newblock In \emph{ICRA}, 2014.

\bibitem[Engel et~al.(2014)Engel, Sch{\"o}ps, and Cremers]{engel2014lsd}
Jakob Engel, Thomas Sch{\"o}ps, and Daniel Cremers.
\newblock Lsd-slam: Large-scale direct monocular slam.
\newblock In \emph{ECCV}, 2014.

\bibitem[Giusti et~al.(2016)Giusti, Guzzi, Cire{\c{s}}an, He, Rodr{\'\i}guez,
  Fontana, Faessler, Forster, Schmidhuber, Di~Caro, et~al.]{giusti2016machine}
Alessandro Giusti, J{\'e}r{\^o}me Guzzi, Dan~C Cire{\c{s}}an, Fang-Lin He,
  Juan~P Rodr{\'\i}guez, Flavio Fontana, Matthias Faessler, Christian Forster,
  J{\"u}rgen Schmidhuber, Gianni Di~Caro, et~al.
\newblock A machine learning approach to visual perception of forest trails for
  mobile robots.
\newblock \emph{IEEE Robotics and Automation Letters}, 2016.

\bibitem[Gupta et~al.(2017)Gupta, Davidson, Levine, Sukthankar, and
  Malik]{gupta2017cognitive}
Saurabh Gupta, James Davidson, Sergey Levine, Rahul Sukthankar, and Jitendra
  Malik.
\newblock Cognitive mapping and planning for visual navigation.
\newblock \emph{arXiv preprint arXiv:1702.03920}, 2017.

\bibitem[Henry et~al.(2012)Henry, Krainin, Herbst, Ren, and
  Fox]{hkhrf-rgbdm-12}
P.~Henry, M.~Krainin, E.~Herbst, X.~Ren, and D.~Fox.
\newblock {RGB-D} mapping: Using kinect-style depth cameras for dense 3d
  modeling of indoor environments.
\newblock \emph{International Journal of Robotics Research}, 2012.

\bibitem[Hoffman et~al.(2014)Hoffman, Guadarrama, Tzeng, Hu, Donahue, Girshick,
  Darrell, and Saenko]{NIPS2014_5418}
Judy Hoffman, Sergio Guadarrama, Eric~S Tzeng, Ronghang Hu, Jeff Donahue, Ross
  Girshick, Trevor Darrell, and Kate Saenko.
\newblock Lsda: Large scale detection through adaptation.
\newblock In Z.~Ghahramani, M.~Welling, C.~Cortes, N.~D. Lawrence, and K.~Q.
  Weinberger, editors, \emph{NIPS}. 2014.

\bibitem[Izadinia et~al.(2017)Izadinia, Shan, and Seitz]{izadinia2017im2cad}
Hamid Izadinia, Qi~Shan, and Steven~M Seitz.
\newblock Im2cad.
\newblock In \emph{CVPR}, 2017.

\bibitem[Kim and Chen(2015)]{kim2015deep}
Dong~Ki Kim and Tsuhan Chen.
\newblock Deep neural network for real-time autonomous indoor navigation.
\newblock \emph{arXiv preprint arXiv:1511.04668}, 2015.

\bibitem[Klein and Murray(2007)]{klein2007parallel}
Georg Klein and David Murray.
\newblock Parallel tracking and mapping for small ar workspaces.
\newblock In \emph{Mixed and Augmented Reality ACM International Symposium on}.
  IEEE, 2007.

\bibitem[Kua et~al.(2012)Kua, Corso, and Zakhor]{kua2012automatic}
John Kua, Nicholas Corso, and Avideh Zakhor.
\newblock Automatic loop closure detection using multiple cameras for 3d indoor
  localization.
\newblock In \emph{IS\&T/SPIE Electronic Imaging}, 2012.

\bibitem[LeCun et~al.(1989)LeCun, Boser, Denker, Henderson, Howard, Hubbard,
  and Jackel]{lecun1989backpropagation}
Yann LeCun, Bernhard Boser, John~S Denker, Donnie Henderson, Richard~E Howard,
  Wayne Hubbard, and Lawrence~D Jackel.
\newblock Backpropagation applied to handwritten zip code recognition.
\newblock \emph{Neural computation}, 1989.

\bibitem[Liu et~al.(2015)Liu, Shen, and Lin]{liu2015deep}
Fayao Liu, Chunhua Shen, and Guosheng Lin.
\newblock Deep convolutional neural fields for depth estimation from a single
  image.
\newblock In \emph{CVPR}, 2015.

\bibitem[Liu et~al.(2016)Liu, Shen, Lin, and Reid]{liu2016learning}
Fayao Liu, Chunhua Shen, Guosheng Lin, and Ian Reid.
\newblock Learning depth from single monocular images using deep convolutional
  neural fields.
\newblock \emph{IEEE transactions on pattern analysis and machine
  intelligence}, 2016.

\bibitem[Michels et~al.(2005)Michels, Saxena, and Ng]{michels2005high}
Jeff Michels, Ashutosh Saxena, and Andrew~Y Ng.
\newblock High speed obstacle avoidance using monocular vision and
  reinforcement learning.
\newblock In \emph{ICML}. ACM, 2005.

\bibitem[Mnih et~al.(2015)Mnih, Kavukcuoglu, Silver, Rusu, Veness, Bellemare,
  Graves, Riedmiller, Fidjeland, Ostrovski, et~al.]{mnih2015human}
Volodymyr Mnih, Koray Kavukcuoglu, David Silver, Andrei~A Rusu, Joel Veness,
  Marc~G Bellemare, Alex Graves, Martin Riedmiller, Andreas~K Fidjeland, Georg
  Ostrovski, et~al.
\newblock Human-level control through deep reinforcement learning.
\newblock \emph{Nature}, 2015.

\bibitem[Mohta et~al.(2014)Mohta, Kumar, and Daniilidis]{mkd-vbcqp-14}
K.~Mohta, V.~Kumar, and K.~Daniilidis.
\newblock Vision based control of a quadrotor for perching on planes and lines.
\newblock In \emph{ICRA}, 2014.

\bibitem[Monajjemi()]{bebob}
Mani Monajjemi.
\newblock Bebop autonomy.
\newblock \url{http://bebop-autonomy.readthedocs.io}.

\bibitem[Pomerleau(1989)]{pomerleau1989alvinn}
Dean~A Pomerleau.
\newblock Alvinn, an autonomous land vehicle in a neural network.
\newblock Technical report, Carnegie Mellon University, Computer Science
  Department, 1989.

\bibitem[Punjani and Abbeel(2015)]{pa-dlhdm-15}
A.~Punjani and P.~Abbeel.
\newblock Deep learning helicopter dynamics models.
\newblock In \emph{ICRA}, 2015.

\bibitem[Puterman and Shin(1978)]{puterman1978modified}
Martin~L Puterman and Moon~Chirl Shin.
\newblock Modified policy iteration algorithms for discounted markov decision
  problems.
\newblock \emph{Management Science}, 1978.

\bibitem[Richter et~al.(2016)Richter, Vineet, Roth, and
  Koltun]{richter2016playing}
Stephan~R Richter, Vibhav Vineet, Stefan Roth, and Vladlen Koltun.
\newblock Playing for data: Ground truth from computer games.
\newblock \emph{arXiv preprint arXiv:1608.02192}, 2016.

\bibitem[Riedmiller(2005)]{m-nfq-05}
Martin Riedmiller.
\newblock Neural fitted q iteration -- first experiences with a data efficient
  neural reinforcement learning method.
\newblock In \emph{European Conference on Machine Learning (ECML)}, 2005.

\bibitem[Ross et~al.(2013)Ross, Melik-Barkhudarov, Shankar, Wendel, Dey,
  Bagnell, and Hebert]{ross2013learning}
St{\'e}phane Ross, Narek Melik-Barkhudarov, Kumar~Shaurya Shankar, Andreas
  Wendel, Debadeepta Dey, J~Andrew Bagnell, and Martial Hebert.
\newblock Learning monocular reactive uav control in cluttered natural
  environments.
\newblock In \emph{ICRA}. IEEE, 2013.

\bibitem[Rusu et~al.(2016)Rusu, Vecerik, Roth{\"o}rl, Heess, Pascanu, and
  Hadsell]{rusu2016sim}
Andrei~A Rusu, Matej Vecerik, Thomas Roth{\"o}rl, Nicolas Heess, Razvan
  Pascanu, and Raia Hadsell.
\newblock Sim-to-real robot learning from pixels with progressive nets.
\newblock \emph{arXiv preprint arXiv:1610.04286}, 2016.

\bibitem[Saxena et~al.(2005)Saxena, Chung, and Ng]{saxena2005learning}
Ashutosh Saxena, Sung~H Chung, and Andrew~Y Ng.
\newblock Learning depth from single monocular images.
\newblock In \emph{NIPS}, 2005.

\bibitem[Schmid et~al.(2013)Schmid, Tomic, Ruess, Hirschmüller, and
  Suppa]{stereo}
Korbinian Schmid, Teodor Tomic, Felix Ruess, Heiko Hirschmüller, and Michael
  Suppa.
\newblock Stereo vision based indoor/outdoor navigation for flying robots.
\newblock In \emph{IROS}, 2013.

\bibitem[Shen et~al.(2013)Shen, Mulgaonkar, Michael, and Kumar]{smmk-vbsea-13}
S.~Shen, Y.~Mulgaonkar, N.~Michael, and V.~Kumar.
\newblock Vision-based state estimation for autonomous rotorcraft mavs in
  complex environments.
\newblock In \emph{ICRA}, 2013.

\bibitem[Siciliano and Khatib(2008)]{handbook_of_robotics}
Bruno Siciliano and Oussama Khatib.
\newblock \emph{Springer handbook of robotics}.
\newblock Springer Science \& Business Media, 2008.

\bibitem[Simonyan and Zisserman(2014)]{VGG16}
Karen Simonyan and Andrew Zisserman.
\newblock Very deep convolutional networks for large-scale image recognition.
\newblock \emph{arXiv preprint arXiv:1409.1556}, 2014.

\bibitem[Tzeng et~al.(2015{\natexlab{a}})Tzeng, Devin, Hoffman, Finn, Peng,
  Levine, Saenko, and Darrell]{tzeng2015towards}
Eric Tzeng, Coline Devin, Judy Hoffman, Chelsea Finn, Xingchao Peng, Sergey
  Levine, Kate Saenko, and Trevor Darrell.
\newblock Towards adapting deep visuomotor representations from simulated to
  real environments.
\newblock \emph{arXiv preprint arXiv:1511.07111}, 2015{\natexlab{a}}.

\bibitem[Tzeng et~al.(2015{\natexlab{b}})Tzeng, Hoffman, Darrell, and
  Saenko]{tzeng2015simultaneous}
Eric Tzeng, Judy Hoffman, Trevor Darrell, and Kate Saenko.
\newblock Simultaneous deep transfer across domains and tasks.
\newblock In \emph{ICCV}, 2015{\natexlab{b}}.

\bibitem[Zhang et~al.(2016)Zhang, Springenberg, Boedecker, and
  Burgard]{zhang2016deeprl}
Jingwei Zhang, Jost~Tobias Springenberg, Joschka Boedecker, and Wolfram
  Burgard.
\newblock Deep reinforcement learning with successor features for navigation
  across similar environments.
\newblock \emph{arXiv preprint arXiv:1612.05533}, 2016.

\bibitem[Zhang(2012)]{zhang2012microsoft}
Zhengyou Zhang.
\newblock Microsoft kinect sensor and its effect.
\newblock \emph{IEEE multimedia}, 2012.

\end{thebibliography}
}

\section{Appendix}
Here we provide several more experiments and ablation study to further evaluate the generalization capability of CAD$^2$RL in unseen environment. Also, we present more details of various experiments of  Section~\ref{experiment} as well as more details of our simulator here.

\section{Synthetic Environment Test}

We conduct experiments on the three test mazes shown in Figure~\ref{fig:training_halls}. 
This experiment is aimed at comparing CAD$^2$RL in the presence of different hallway geometries, distractors, and obstacles, in synthetic hallways that are distinct from the ones used but similar in visual style. Note that the test hallways were intentionally designed to be larger and more challenging. We rendered all images at the test time using a randomization of $100$ different test textures that were not seen at the training time. We tested our method and the prior baseline methods in two conditions: in the first condition, the hallways contained randomly placed furniture, and in the second, no furniture was present. Both scenarios have fixtures such as open or closed doors, windows, and paintings, but the hallways with furniture provide an additional challenge due to the more complex geometry, which is substantially more elaborate than the scanned hallway used in the previous section.

\subsection{Evaluation criteria}
\label{sec:synth_exp}
\begin{figure*}[t]
    \centering
    \includegraphics[width=1\textwidth]{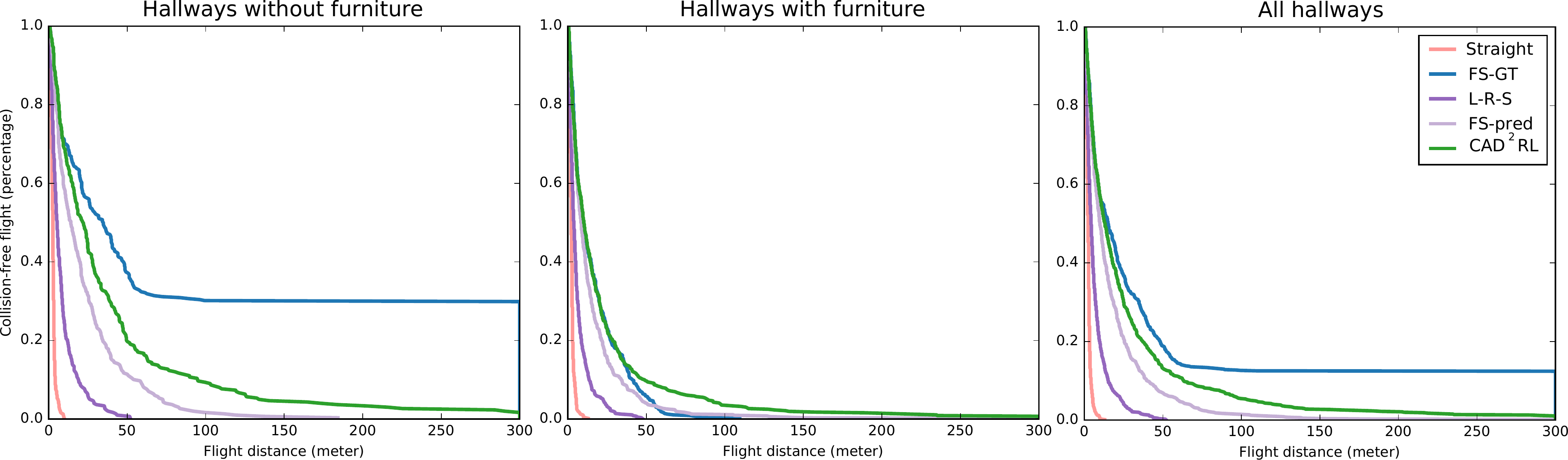}
    	\vspace{-0.15in}
    \caption{Quantitative results on the simulated test hallways. Aside from the upper bound baseline FS-GT, which uses ground truth collision raycasts, our method CAD$^2$RL, achieves the longest collision-free flights.}
    \label{sim_quantitative}
    \vspace{-0.15in}
\end{figure*}
Our aim is to evaluate the performance of our trained policy in terms of the duration of collision free flight. To do this, we run continuous episodes that terminate upon experiencing a collision, and count how many steps are taken before a collision takes place. We set the maximum number of steps to a fixed number throughout each experiment. 

An episode can begin in any location in the choice environment, but the choice of the initial position can have a high impact on the performance of the controller, since some parts of the hallway, such as dead ends, sharp turns, or doorways, can be substantially more difficult. Therefore, to make an unbiased evaluation and to test the robustness of the learned policies, we start each flight from a location chosen uniformly at random within the free space of each environment. We use random initialization points and keep them fixed throughout all experiments to provide a fair comparison between different methods, including prior work. In the experiments, the quadrotor has constant velocity during each flight, and we convert the number of steps to meters in order compute the distance traveled in each flight.

We evaluate performance in terms of the percentage of trials that reached a particular flight length. To that end, we report the results using a curve that plots the distance traveled along the horizontal axis, and the percentage of trials that reached that distance before a collision on the vertical axis. More formally, vertical axis represent $\sum_{i=1}^{|T|}{(1-\mathbbm{1}(C_{d}(T_{i})))}$ where $T$ denotes a trial and $C_{d}(T_{i})$ is the event of collision happening for the $i$th trail at distance $d$ in the x axis. This provides an accurate and rigorous evaluation of each policy, and allows us to interpret for each method whether it is prone to collide early in the flight, or can maintain collision-free flight at length. Note that achieving completely collision-free flight in all cases from completely randomized initial configurations is exceptionally difficult. Please note that we used the same evaluation metric in the quantitative experiment in section~\ref{sec:real_exp}.

We compare the performance of our method with previous methods for learning-based visual obstacle avoidance introduce in Section~\ref{sec:real_exp}. In addition, we compare against \emph{ground truth free space controller} baseline which simulates a quadrotor equipped with perfect LIDAR range sensors or depth cameras. The performance of this baseline shows the upper-bound of the performance of a free-space prediction based controller. The policy always selects the most central free-space labeled bin in the spatial grid of the current image. Note that this does not always result in the best performance, since the behavior of this baseline is myopic, but it does serve to illustrate the difficulty of the test environments.

\subsection{Results and Analysis}
We randomly sampled $100$ random locations as initialization point. These points were selected uniformly from challenging locations such as junctions as well as less challenging locations in the middle of hallways. The velocity was $0.3$ meters per step.
Figure~\ref{sim_quantitative} compares the performance of our method compared with other baselines in each of the ``with furniture" and ``without furniture" test scenarios. CAD$^2$RL consistently outperforms the other baselines, as well as the model trained with supervised learning for free space prediction, FS-pred. In the hallways that do not contain any furniture, the ground truth free space baseline (FS-GT) obtains the best performance, while the presence of furniture in the second test scenario effectively reduce its performance due to the greedy strategy. In both scenarios, CAD$^2$RL has the highest performance among the learning-based methods.

\section{Free Space Prediction Evaluation}
In this experiment, we are interested to see how well our free space prediction model can detect free spaces and obstacles compared with its performance on the synthetic images. To this end, we used the simulator to compute the mask of Free Space (FS)/Obstacle(O) of $4$k rendered frames which were sampled uniformly along the hallways. For the performance metrics, we use precision and Jaccard similarity. The precision shows the ratio of correctly labeled pixels as corresponding to ``FS'' or ``O''. The Jaccard similarity is the intersection over union of the result and ground truth labels for both free-space and obstacle labels.
Table~\ref{tab:frespace_pred} summarizes the obtained results. The first row of the table shows the results obtained for the images rendered in from our test hallways with test textures. The second row shows the results obtained on the photo realistic images of~\cite{kua2012automatic}. Although, there is a $10\%$ precision gap between the performance on synthetic images and photo-realistic images, which is due to the domain shift, the obtained precision results on the unseen photo-realistic images is high, i.e. $80\%$.
Note that our synthetic hallways are much narrower than the real hallways of~\cite{kua2012automatic}. This results in smaller free-space areas and larger obstacle areas in the synthetic images compared with ~\cite{kua2012automatic} where images have more balanced distribution of free-space vs obstacles. This results in lower Jaccard(FS) in the synthetic images.

\begin{table}
\begin{small}
\begin{center}
\caption{\small Pixel-wise free space prediction results.}
\begin{tabular}{lccc} 
\toprule
 & Precision & Jaccard(FS) & Jaccard(O)  \\
\midrule
Synthetic hallways & { 90.38} & 56.99 & { 87.26}  \\
Realistic hallways& { 80.33} & 63.39 & { 63.30}  \\
\bottomrule
\end{tabular}
\end{center}
\end{small}
\vspace{-0.2in}
\label{tab:frespace_pred}
\end{table}

\section{Details on real environment tests}
We used the Parrot Bebop drone controlled via the ROS Bebop autonomy package~\cite{bebob}. We ran real flight experiments using two different drone platforms: the Parrot Bebop 1.0 and the Bebop 2.0. Although these two drones have similar SDK, they have different physical specifications in terms of dimensions, weight, and maximum speed. Note that the images produced by the onboard drone camera are center cropped to remove the fish-eye effect, and thus the objects appear closer than they really are. 

Figure~\ref{real_flight_samples} shows the sequence of images captured by the flying drone in various scenarios (for the more complete sequences please check the supplementary video \href{https://youtu.be/nXBWmzFrj5s}{https://youtu.be/nXBWmzFrj5s}). The red dots show the action computed by our policy. Our controller can successfully navigated the drone throughout free spaces while avoiding collision. Due to imperfect stabilization, turbulence and air currents, the drone may sometimes drift to the left or right, and our controller can recover from these situations by stabilizing the drone at each time step based on the current image observation. This suggests that, though our model is fully trained in simulation without seeing any real images or using any human provided demonstration, it has the capability to generalize to real-world images and conditions. In the following sections we evaluate the real flight performance both quantitatively and qualitatively.

\section{Details about our Simulator Setup}
We designed a total of $24$ different hallways using $12$ different floorplans. Some of our hallways are populated with furnitures while some of them are not.
We use $21$ different items of furniture that commonly exist in the hallways (benches, chairs, etc). We have slightly randomized the size of these furniture to provide diversity. The walls are textured with randomly chosen textures, chosen from a pool of 200 possible textures (e.g. wood, metal, textile, carpet, stone, glass, etc.), and illuminated with lights that are placed and oriented at random. Pretraining images are generated by flying a simulated camera through the hallways with randomized height and random perturbations of the yaw angle, in order to provide a diversity of viewpoints. In all, we randomize textures, lighting, furniture placement (including placement and identity of furniture items), and camera position and angle.

\end{document}